\documentclass{article}

\usepackage{spconf}
\usepackage{soul}
\usepackage{hyperref}
\usepackage{subcaption}
\usepackage{booktabs}
\usepackage{amsmath,amssymb,amsfonts}
\usepackage{algorithmic}
\usepackage{graphicx}
\usepackage{textcomp}
\usepackage{xcolor}
\def\BibTeX{{\rm B\kern-.05em{\sc i\kern-.025em b}\kern-.08em
    T\kern-.1667em\lower.7ex\hbox{E}\kern-.125emX}}
\usepackage{enumitem}

\begin{document}


\title{
RETHINKING HUMAN-ALIGNED EVALUATION: AN ANALYSIS OF SEMANTIC METRICS BEYOND WER
}

\name{%
  \begin{tabular}{c}
    Hritika Sharma$^{*1}$, 
    Thibault Baneras-Roux$^{*2}$, 
    Alessandra Pinto$^{3}$, 
    Petr Motlicek$^{2}$, \\
    Hyunggu Jung$^{4}$, 
    Esaú Villatoro-Tello$^{2}$, 
    Somang Nam$^{\star}$
  \end{tabular}%
  \thanks{* Equal contribution.}%
}


\address{
    $^{1}$ Faculty of Computer Science and Technology, Algoma University, $^{2}$ Idiap Research Institute, \\
    $^{3}$ School of Psychology, Algoma University, $^{4}$ School of Nursing, Seoul National University
}

\maketitle

\begin{abstract}
Word Error Rate (WER), the most commonly used metric for Automatic Speech Recognition (ASR), treats every lexical deviation from the reference as equally costly, regardless of whether it changes meaning. This raises the question: does WER actually track how humans judge ASR transcript quality? We introduce HATS-en, an English dataset for human-centered ASR evaluation. Using this dataset, we benchmark lexical metrics against several configurations of BERTScore and SemDist, varying the language model, layer, and pooling strategy. We find that WER agrees least with human judgment among all metrics tested, that the best-performing SemDist configurations achieve the highest overall agreement, ahead of CER and BERTScore, and that no single model is best across settings. CER, despite its simplicity and low cost, remains remarkably close to these best configurations.
In line with prior recommendations, our results support shifting ASR evaluation toward CER both for English and for morphosyllabic writing systems as it is a more interpretable and low-cost metric for what evaluation should actually capture, and using SemDist as a complementary evaluation.
\end{abstract}
\begin{keywords}
ASR, Human Perception, Evaluation Metrics, Benchmark Dataset
\end{keywords}



\section{Introduction}
\label{sec:intro}
Automatic Speech Recognition (ASR) has become an important component of human-computer interaction, supporting applications such as virtual assistants, automatic captioning, and real-time meeting transcription.
Recent advances in self-supervised learning and transformer-based architectures have significantly improved ASR performance~\cite{wav2vec2,hubert,whisper}.
Evaluating ASR systems is essential both to diagnose a system's weaknesses and to compare competing systems for deployment, but both purposes assume the metric used faithfully reflects transcript quality as perceived by end users.
While this assumption is not always warranted~\cite{favre2013automatic, semdist2022, kafle}, ASR systems are traditionally assessed using lexical metrics such as WER, which remains the most widely adopted metric in the field.

\begin{figure}[t]
    \centering
    \includegraphics[width=\columnwidth]{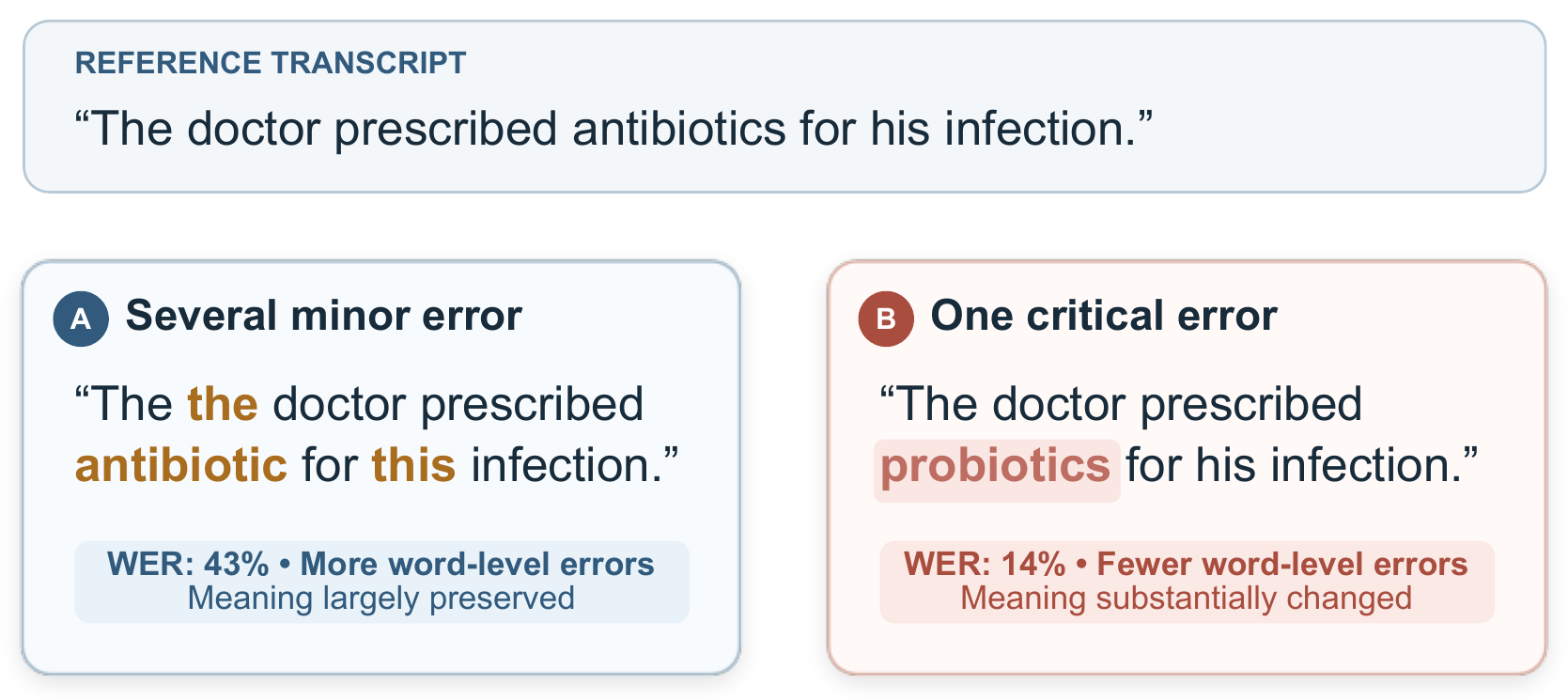}
    \caption{Illustration of the divergence between WER and semantic fidelity.}
    \label{fig:example}
\end{figure}

WER is simple, interpretable, and easy to compute, but it measures differences only at the word level.
It may not fully capture changes in meaning or perceived transcript quality, as two transcripts with comparable WER can differ in how a human reader would judge them (see Figure~\ref{fig:example}).
Semantic metrics~\cite{semdist2022, bertscore} were developed to address this gap, but their validation requires large-scale human judgments. To our knowledge, no openly accessible English-language resource exists at sufficient scale.

In this work, we address this gap along two complementary directions. 
First, we introduce \textbf{HATS-en}, an English extension of the HATS~\cite{hats} side-by-side annotation protocol, derived from the LibriSpeech test-clean set and comprising 1,000 human-annotated transcript comparison triplets generated using four ASR systems. 
Second, we use HATS-en to conduct a large-scale comparison of automatic evaluation metrics against human preference, covering WER, CER, standard semantic metrics and a systematic sweep of semantic metric configurations (language model, layer, and pooling strategy) for BERTScore and SemDist. 
The complete dataset, including audio segment identifiers, reference transcripts, ASR hypotheses, and the distribution of systems across pairs, is publicly \href{https://anonymous.4open.science/r/HATS-en/hats-en-systems.csv}{available} in order to support future development of ASR metrics that more closely reflect human judgments.

\section{Related Work}
\label{sec:prior}

\subsection{ASR Systems and Evaluation Metrics}

Modern ASR systems differ in their design choices, ranging from self-supervised approaches such as wav2vec 2.0~\cite{wav2vec2} and HuBERT~\cite{hubert} to large-scale weakly supervised models such as Whisper~\cite{whisper}.
Systems vary in tokenization strategy, in whether they incorporate an external or integrated language model, and in the use of rescoring or decoding heuristics~\cite{udagawa22b}.
These architectural differences affect overall accuracy and the produced error types, which can lead to different rankings per evaluation metric \cite{baneras2024comprehensive}.

Lexical metrics penalize any surface deviation from the reference regardless of whether it changes meaning, which can cause transcripts with similar error rates to differ in perceived quality~\cite{semdist2022}. 
Work examining the impact of transcription errors on downstream tasks has similarly shown that not all errors are equally harmful~\cite{munteanu2006, favre2013automatic}.

In response, the community has proposed several semantic evaluation metrics.
BERTScore compares reference and hypothesis texts using contextual token embeddings~\cite{bertscore}; SemDist enables efficient comparison of sentence-level meaning through cosine similarity~\cite{semdist2022} and can also use Sentence-BERT embeddings~\cite{sentencebert}.
Task-oriented metrics such as Meaning Error Rate (MER) go further by weighting errors according to their impact on domain-specific meaning rather than treating all lexical substitutions equally~\cite{mer}.
These metrics differ in formulation and in configuration choices whose effect on agreement with human judgment has not been systematically studied.

\subsection{Human-centered Evaluation Benchmarks}

Human-centered studies have underlined that transcript quality, as perceived by end users, is not always well predicted by lexical accuracy~\cite{semdist2022} and that speech recognition accuracy alone is insufficient to measure caption quality~\cite{kafle, nam2023developing}.

To measure this disconnect, a resource was built in French, 
allowing to measure agreement between metrics and human preference~\cite{hats}. 
This comparison showed that several semantic metrics, particularly using Sentence-BERT level embeddings and BERTScore, achieved greater agreement with human judgments than WER~\cite{hats}.
Subsequent work on this French resource further showed that large language models (LLM) can be prompted to predict human preferences over ASR hypotheses~\cite{baneras2026evaluation}.
Despite this progress, existing English resources remain limited.
To our knowledge, the only comparable effort is a small dataset of 50 utterances~\cite{thennal2025advocating}, relying on direct assessment of a single score for each reference-hypothesis pair. 

However, this design is vulnerable to anchor bias: an annotator's internal notion of what merits a given score shifts over the course of a session and is influenced by previously seen transcripts.
As a result, two annotators, or even the same annotator at different points in a session, may assign different scores to equivalent transcripts.
Pairwise comparison reduces this issue, since both hypotheses are judged relative to one another.
HATS-en, together with the metric analyses presented in this paper, is designed to address this gap.

\begin{table}[t]
\centering
\caption{WER and CER (\%) of the four ASR systems on LibriSpeech test-clean.}
\label{tab:asr-performance}
\begin{tabular}{lcc}
\toprule
Model & WER (\%) & CER (\%) \\
\midrule
GoogleSR        & 8.22 & 4.12 \\
Whisper-base    & 5.17 & 1.93 \\
Wav2vec-FT-ASR  & 3.28 & 0.97 \\
CRDNN-RNNLM     & 3.10 & 1.26 \\
\bottomrule
\end{tabular}
\end{table}

\section{The HATS-en Dataset}
\label{sec:dataset}

\subsection{Annotation Protocol}

HATS-en extends the Human-Assessed Transcription Side-by-side (HATS) protocol~\cite{hats} to English.
For each audio segment, annotators are shown: a reference transcript alongside two ASR hypotheses and asked to choose the one that better preserves meaning, reads more fluently, and is easier to understand.
Audio is deliberately withheld, so that judgments are grounded in transcript quality alone.

\begin{figure}[htbp]
    \centering
    \begin{subfigure}[b]{0.47\columnwidth}
        \centering
        \includegraphics[width=\linewidth]{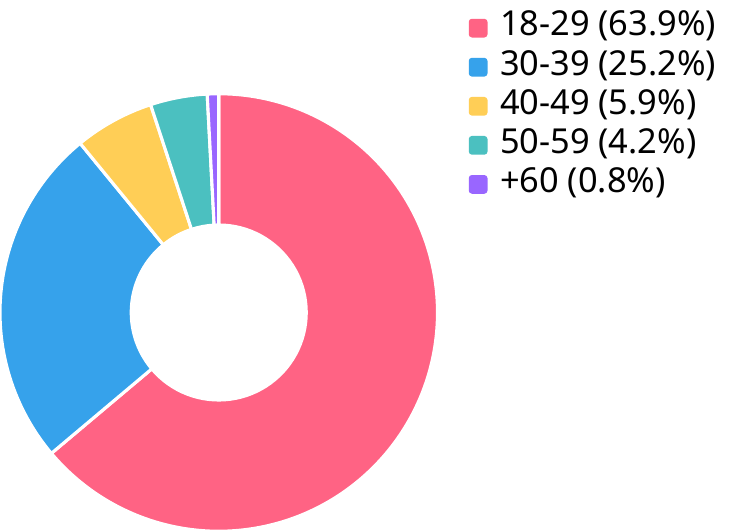}
        \caption{Age distribution}
        \label{fig:age}
    \end{subfigure}
    \hfill
    \begin{subfigure}[b]{0.47\columnwidth}
        \centering
        \includegraphics[width=\linewidth]{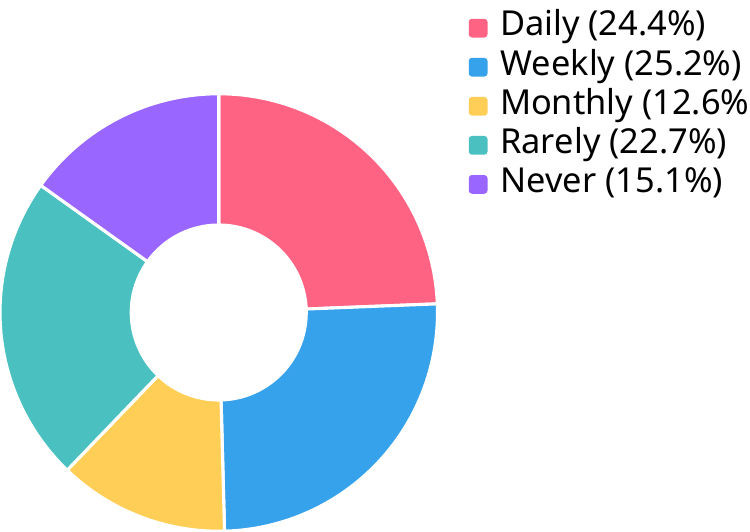}
        \caption{ASR usage}
        \label{fig:asr}
    \end{subfigure}
    \vspace{0.5cm}
    \begin{subfigure}[b]{0.47\columnwidth}
        \centering
        \includegraphics[width=\linewidth]{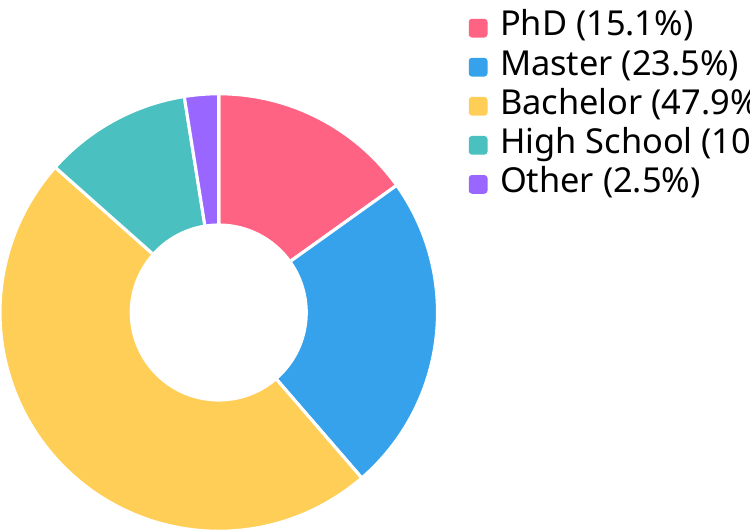}
        \caption{Education}
        \label{fig:education}
    \end{subfigure}
    \hfill
    \begin{subfigure}[b]{0.47\columnwidth}
        \centering
        \includegraphics[width=\linewidth]{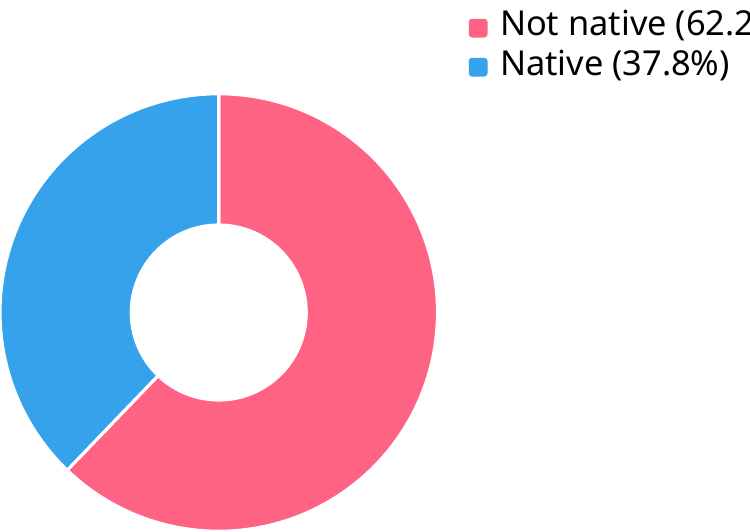}
        \caption{Native language}
        \label{fig:native}
    \end{subfigure}
    \caption{Demographic characteristics of annotators.}
    \label{fig:demographics}
\end{figure}

Annotation was carried out through a web application presenting each reference and its hypothesis pair side by side, with hypothesis order randomized to control for positional bias and forced-choice selection to prevent ties.
The 1,000 comparison instances were split into 50 batches of 20 triplets each, and annotators were assigned one or more batches to complete.
A total of 119 annotators provided valid completed annotations across two sequential campaigns: 1) volunteer annotators contributing without compensation, and 2) with a lottery incentive of a one-sixth chance to win \$20 CAD.

\subsection{Speech Data and ASR Systems}

Transcripts are drawn from the test-clean subset (2,620 segments, $\approx$5 hours of speech) of LibriSpeech~\cite{librispeech}, a corpus of read English audiobook speech.
We used test-clean to reduce potential bias from models trained on the training set of Librispeech.
Each segment was transcribed by four systems chosen to span distinct design paradigms: \href{https://pypi.org/project/SpeechRecognition/}{GoogleSR}, using a speech-to-text API; \href{https://huggingface.co/facebook/wav2vec2-base-960h}{Wav2vec-FT-ASR}, a fine-tuned CTC model based on Wav2vec2~\cite{wav2vec2}; \href{https://huggingface.co/speechbrain/asr-crdnn-rnnlm-librispeech}{CRDNN-RNNLM}, a hybrid CRDNN acoustic model with RNN language model rescoring trained with SpeechBrain~\cite{speechbrain}; and \href{https://huggingface.co/openai/whisper-base}{Whisper-base}~\cite{whisper}.

Table~\ref{tab:asr-performance} reports the WER and CER of each system on the test-clean set, confirming that the four systems differ in raw transcription accuracy in addition to architecture. We also note that the system hierarchy differs between WER and CER.
We retain only hypothesis pairs in which at least one transcript differs from the reference, since identical transcripts provide no basis for a preference judgment.
From this pool, we randomly sample 1,000 comparison instances, each pairing a reference transcript with two distinct ASR hypotheses.
The randomized hypothesis order mitigates the bias of position being the annotator choice, yet we nonetheless observe a small bias (5\%) toward selecting hypothesis A when the two hypotheses are close in quality.

\section{Agreement with Human Judgments}
\label{sec:agreement}
We evaluate how automatic metrics reproduce human preferences on the HATS-en dataset.
For each metric configuration, we compute a score between the reference and each hypothesis in order to rank the hypotheses; then, we measure agreement as the percentage of instances for which the metric selects the hypothesis preferred by the human annotators.
Annotations without unanimous agreement among annotators were discarded to reduce ambiguity.

To assess the reliability of the collected judgments, we compute Fleiss' Kappa over triplets with annotations. 
We obtain a Kappa of 0.3061 ($p < 10^{-60}$), indicating fair agreement beyond chance.
Annotators reached unanimous agreement on 48.10\% of triplets, compared to the 25\% expected under random annotation.
This suggests that some hypothesis pairs are difficult to distinguish, motivating our consensus filtering: only triplets with unanimous agreement are retained for metric evaluation.
The metric evaluation therefore focuses on instances with unanimous human preference.
As reference baselines, conventional WER and CER achieve agreement rates of 75.47\% and 86.28\%, respectively, after applying the same consensus filtering. This difference is statistically significant (McNemar's exact test, $p = 9.03 \times 10^{-12}$), confirming that character-level granularity provides a measurably better correspondence with human judgments than word-level edit distance on this dataset.
\begin{figure*}[htbp]
    \centering
    \begin{subfigure}[b]{0.32\textwidth}
        \centering
        \includegraphics[width=\linewidth]{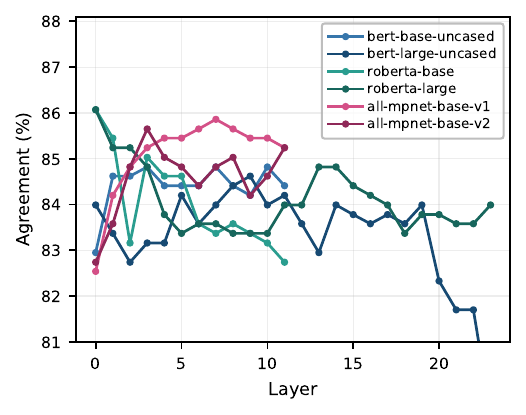}
        \caption{BERTScore: Encoder-based models}
        \label{fig:bertscore-bertlike}
    \end{subfigure}
    \hfill
    \begin{subfigure}[b]{0.32\textwidth}
        \centering
        \includegraphics[width=\linewidth]{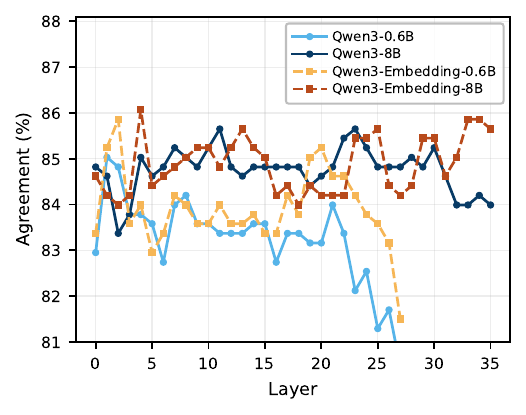}
        \caption{BERTScore: Qwen models}
        \label{fig:bertscore-qwen}
    \end{subfigure}
    \hfill
    \begin{subfigure}[b]{0.32\textwidth}
        \centering
        \includegraphics[width=\linewidth]{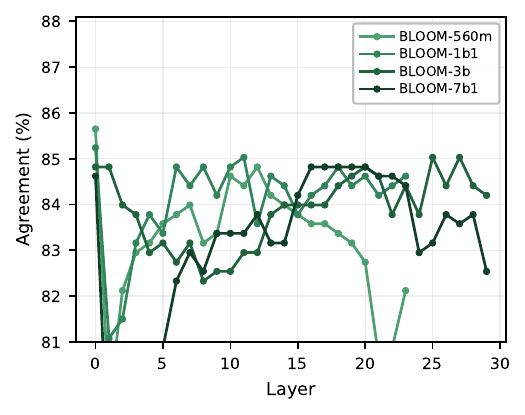}
        \caption{BERTScore: BLOOM models}
        \label{fig:bertscore-other-causal}
    \end{subfigure}
    \centering
    \begin{subfigure}[b]{0.32\textwidth}
        \centering
        \includegraphics[width=\linewidth]{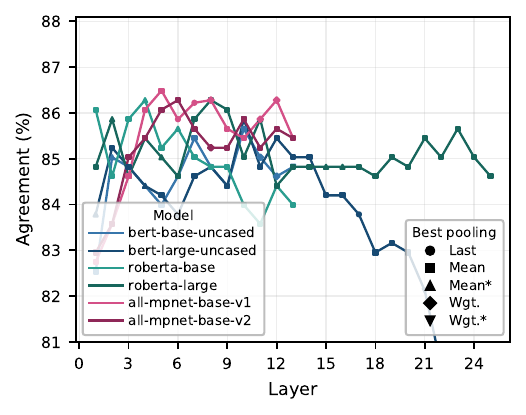}
        \caption{SemDist: Encoder-based models}
        \label{fig:semdist-bertlike}
    \end{subfigure}
    \hfill
    \begin{subfigure}[b]{0.32\textwidth}
        \centering
        \includegraphics[width=\linewidth]{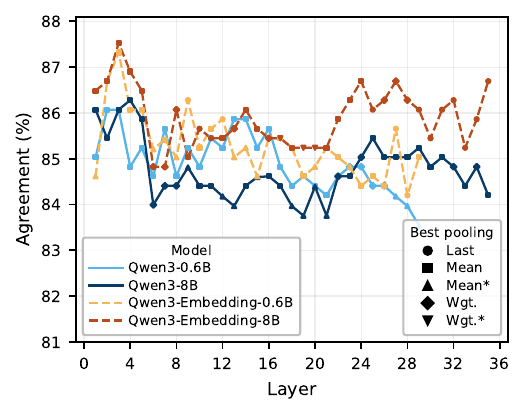}
        \caption{SemDist: Qwen models}
        \label{fig:semdist-qwen}
    \end{subfigure}
    \hfill
    \begin{subfigure}[b]{0.32\textwidth}
        \centering
        \includegraphics[width=\linewidth]{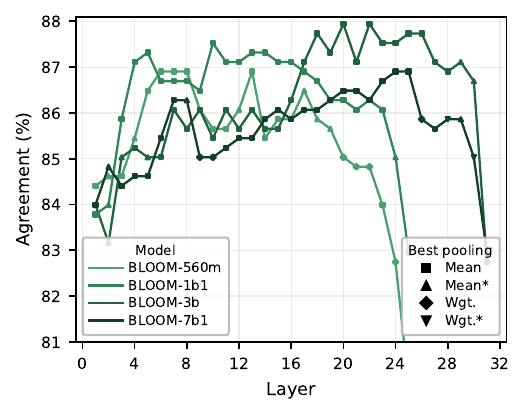}
        \caption{SemDist: BLOOM models}
        \label{fig:semdist-other-causal}
    \end{subfigure}
    \caption{Agreement (\%) of BERTScore and SemDist with human perception on HATS-en. Metrics are evaluated using different LLMs, layers, and pooling strategies; only the best-performing pooling strategy is reported.}
\label{fig:agreement}
\end{figure*}

\subsection{BERTScore}

We first evaluate BERTScore using representations extracted from different layers of several language models (see Figure~\ref{fig:agreement} for the human agreement comparison of BERTScore).
The results reveal variation across layers, demonstrating that the representation used by BERTScore has a strong influence on its agreement with humans.
In particular, the best-performing layers are not consistently the final layers, where several models achieve their highest agreement at intermediate depths.
This suggests that representations tied to the final language modeling objective do not always match how humans subjectively judge ASR quality.
Models with similar architectures exhibit similar layer-wise patterns; for instance, the embedding-oriented variants of Qwen models follows slightly better but comparable trends.

Despite requiring substantially more computation, the best BERTScore outperform WER but do not exceed CER's agreement, which remains a particularly strong baseline.

\subsection{SemDist}

The semantic similarity was evaluated using SemDist, a distance derived from the cosine similarity between sentence-level embeddings obtained by aggregating token representations from LLMs.
This setup allows us to study the language model, representation layer, and how token-level representations should be pooled into a sentence-level embedding (see Figure~\ref{fig:agreement} for the human agreement comparison of SemDist).
Given a sequence of token embeddings produced by an LLM, a fixed-size sentence representation is obtained via pooling.
We evaluate multiple strategies, where $t_i$ denotes the embedding of token $i$ in a sequence of length $n$:
\begin{itemize}[nosep]
\item \textbf{Last token (\textit{Last})}: $t_n$
\item \textbf{Mean (\textit{Mean})}: $\tfrac{1}{n}\sum_{i=1}^{n} t_i$
\item \textbf{Mean without last (\textit{Mean*})}: $\tfrac{1}{n-1}\sum_{i=1}^{n-1} t_i$
\item \textbf{Weighted mean (\textit{Wgt.})}: $\tfrac{\sum_{i=1}^{n} i \times t_i}{\sum_{i=1}^{n} i}$
\item \textbf{Weighted mean without last (\textit{Wgt.*})}: $\tfrac{\sum_{i=1}^{n-1} i \times t_i}{\sum_{i=1}^{n-1} i}$
\end{itemize}

SemDist achieves higher agreement with human preferences than the corresponding BERTScore configurations, indicating that sentence-level semantic representations can provide a stronger signal for distinguishing between ASR hypotheses.
As with BERTScore, models with similar architectures exhibit similar patterns across layers.
The Qwen models and their embedding-oriented counterparts show similar behavior, although the embedding-oriented models generally achieve better agreement, suggesting that their representations align more closely with human judgments.

Mean-based pooling strategies perform strongest overall, with several best-performing configurations using variants that exclude the last token (Mean*) or apply position-based weighting (Wgt. and Wgt.*).
In contrast, relying solely on the last token seems to be less competitive, except for embedding-oriented Qwen models toward the last layers, possibly reflecting their embedding-oriented fine-tuning.
This highlights that aggregating information across the sequence is important for obtaining sentence representations that align with human judgments of ASR quality.
Several SemDist configurations achieve agreement above the CER baseline.
In particular, the Qwen embedding models and the BLOOM models reach agreement rates higher than CER for their strongest configurations.
The best-performing configuration, SemDist$_{\text{Bloom-3b}}$, significantly outperforms WER (McNemar's exact test, $p = 5.87 \times 10^{-15}$).
Its advantage over CER, however, does not reach statistical significance on the consensus-filtered subset ($p = 0.243$), although the trend is directionally consistent and approaches significance on the full, unfiltered corpus ($p = 0.052$).
Similarly, the comparison between SemDist$_{\text{Bloom-3b}}$ and the best-performing BERTScore configuration (Qwen-based) shows a directional but non-significant advantage ($p = 0.136$).
Similarly sized models can exhibit different performance depending on their architecture, yet larger models often achieve better agreement.
This suggests that the scale can be advantageous with an appropriate representation and pooling strategy.
Overall, the results emphasize that no single parameter determines metric quality because the correspondence with human perception depends on LLM, layer, and pooling strategy.


\section{Conclusion}
\label{sec:conclusion}
Our results indicate that WER, despite its systematic use, is not the metric that best tracks human preference among ASR hypotheses.
We therefore argue that the community should report CER by default, with a validated semantic metric as a secondary check, including for English and other non-morphosyllabic writing systems where CER has been treated as a niche measure~\cite{thennal2025advocating}.
Future work should extend the dataset to spontaneous speech, additional ASR systems and domains, and examine the cases where automatic metrics disagree with human judgments.
Building on recent findings showing that LLMs can predict human preferences for ASR outputs, future work should evaluate LLM-based judges against conventional and semantic metrics ~\cite{baneras2026evaluation}. 
Finally, the relationship between human preference, automatic metrics, and downstream task performance could be investigated to understand which dimensions of ASR quality are most relevant for practical applications.

\section{Acknowledgments}
This work was supported by the Idiap Research Institute and the Algoma University Research Fund. We thank all annotators for their careful and dedicated work throughout the annotation process.

\section{Compliance with Ethical Standards}
The data collection protocol used in this research was reviewed and approved by the Research Ethics Board (REB) of Algoma University.





\bibliographystyle{IEEEbib}
\bibliography{refs}

\end{document}